# Element-wise Multiplication Based Deeper Physics-Informed Neural Networks

Feilong Jiang[a] Xiaonan Hou[a]* Min Xia[b]*
[a] Department of Engineering, Lancaster University, LA1 4YW Lancaster, U.K.
[b] Department of Mechanical and Materials Engineering, University of Western Ontario, London, Ontario, Canada
x.hou2@lancaster.ac.uk
mxia47@uwo.ca

## Abstract

As a promising framework for resolving partial differential equations (PDEs), Physics-Informed Neural Networks (PINNs) have received widespread attention from industrial and scientific fields. However, lack of expressive ability and initialization pathology issues are found to prevent the application of PINNs in complex PDEs. In this work, we propose Deeper Physics-Informed Neural Network (Deeper-PINN) to resolve these issues. The element-wise multiplication operation is adopted to transform features into high-dimensional, non-linear spaces. Benefiting from element-wise multiplication operation, Deeper-PINNs can alleviate the initialization pathologies of PINNs and enhance the expressive capability of PINNs. The proposed structure is verified on various benchmarks. The results show that Deeper-PINNs can effectively resolve the initialization pathology and exhibit strong expressive ability.

## 1. Introduction

As an important role in scientific machine learning field, Physics-Informed Neural Networks (PINNs) appear to be a promising method for resolving partial differential equations (PDEs) [1]. Utilizing the physical laws as the constraints of neural networks, PINNs show better interpretability and generalization ability than traditional data-driven machine learning models. As an alternative of traditional numerical algorithms, PINNs have been widely applied in scientific and engineering fields [2-5].

Although promising outcomes have been presented, PINNs have been observed of failing resolving some PDEs, especially when the solutions exhibit high-frequency or multi-scale characteristics [6], [7]. Many efforts have been made to improve the capability of PINNs. To resolve the loss terms unbalancing problem, methods such as NTK-based weighting [6], Grad Norm [8] and SoftAdapt [9] are used to weight different loss term for training PINNs effectively. Utilizing the geometry-aware approximations [10] or Fourier feature embedding [11] methods, the boundary conditions can be exactly imposed during the training process. Consequently, the accuracy of PINNs can be effectively improved. Since the training process of PINNs may go against physical causality, training methods that follow the spatio-temporal causal are proposed to improve the accuracy of PINNs [12-14]. The loss imbalance problem between different training points prevents PINNs from resolving PDEs correctly. This issue can be resolve by introducing adaptive weighting scheme [15-18] or adaptive resampling method [19], [20]. New neural network structures such as mMLP [8], Fourier feature embedding [7], DM-PINNs [21] and SPINN [22] are also proven to be useful for improving the representative capability of PINNs.

Despite the success of these research, most existing works still use small neural networks with a few layers. This is attributed to the gradient vanishing issue of deep PINNs, which limits the representative capability of PINNs. Recent research has proofed that the initialization pathologies [23] of PINNs can intensifies vanishing gradient, which prevents the PINNs from going deeper. To overcome this limitation, we propose a novel architecture which is Deeper-PINN. By doing element-wise multiplication between different sub-layers, the outputs of these layers are fused together and projected into nonlinear-high-dimensional feature spaces. In this way, Deeper-PINNs can still keep nonlinear expressive ability even all the layers of the neural network operate within a linear regime. Consequently, the initialization pathologies can be eliminated. Besides, the element-wise multiplication operations enable the Deeper-PINNs with stronger expressive capability, which means the Deeper-PINNs have better parameter efficiency. With the stacking of a series of shortcut connected element-wise multiplication blocks, the model can use deeper neural network structures and obtain better results.

The rest of the paper is organised as follows. In Section 2, we give a brief introduction of PINNs and its initialization pathologies. In Section 3, we present the structure of Deeper-PINNs, and theoretically and empirically proof the effectiveness of Deeper-PINNs for alleviating the initialization pathologies. Section 4 presents the verifications of Deeper-PINNs on several benchmarks. Finally, in Section 5, we summarize with a discussion of our work.

## 2. Related works

**Physics-informed Neural Networks (PINNs)**

Following the original work of physics-informed neural networks (PINNs) in [1], the typical form of partial differential equation (PDEs) can be represented as:

$$u_t + \mathcal{N}\left[u\right] = 0,\ t \in \left[0,T\right],\ x \in \Omega, \tag{1}$$

where $u$ indicates the latent solution, $\mathcal{N}\left[\cdot\right]$ represents a linear or nonlinear differential operator. The initial and boundary conditions are in the forms of:

$$u\left(0,x\right) = g\left(x\right),\ x \in \Omega, \tag{2}$$

$$\mathcal{B}\left[u\right] = 0,\ t \in \left[0,T\right],\ x \in \partial\Omega, \tag{3}$$

where $g\left(x\right)$ is a given function, $\mathcal{B}\left[\cdot\right]$ indicates boundary operator.

The core idea of PINNs is approximating the latent solution $u\left(t,x\right)$ with a neural network $u_\theta\left(t,x\right)$, where $\theta$ represents trainable parameters of the neural network. Instead of utilizing the difference between label data $u\left(t,x\right)$ and the prediction $u_\theta\left(t,x\right)$, PINNs directly utilize the PDE and the corresponding initial and boundary conditions as the loss function. Utilizing automatic differentiation [24], the required gradients with respect to input variables can be obtained. The loss function can be expressed as:

$$\mathcal{L}\left(\theta\right) = \lambda_{ic}\mathcal{L}_{ic}\left(\theta\right) + \lambda_{bc}\mathcal{L}_{bc}\left(\theta\right) + \lambda_r\mathcal{L}_r\left(\theta\right), \tag{4}$$

where

$$\mathcal{L}_{ic}\left(\theta\right) = \frac{1}{N_{ic}}\sum_{i=1}^{N_{ic}}\left|u_\theta\left(0,x_{ic}^i\right) - g\left(x_{ic}^i\right)\right|^2, \tag{5}$$

$$\mathcal{L}_{bc}\left(\theta\right) = \frac{1}{N_{bc}}\sum_{i=1}^{N_{bc}}\left|\mathcal{B}\left[u_\theta\right]\left(t_{bc}^i, x_{bc}^i\right)\right|^2, \tag{6}$$

$$\mathcal{L}_r\left(\theta\right) = \frac{1}{N_r}\sum_{i=1}^{N_r}\left|\frac{\partial u_\theta}{\partial t}\left(x_r^i, t_r^i\right) + \mathcal{N}\left[u_\theta\right]\left(x_r^i, t_r^i\right)\right|^2, \tag{7}$$

$\lambda$ represents the weighting coefficient, which is used to balance different loss terms [25].

Under these conditions, the training procedure of PINNs can be interpreted as solving the given PDEs.

**Tackling initialization pathologies of PINNs**

For PINNs, the derivative operations of the output $u_\theta$ is implemented before the backpropagation. As proven by Wang et al. [23], such operations will reduce the nonlinear expressively ability of PINNs at the initialization state especially when the number of layers increases. Consequently, the initialization pathologies intensify the vanishing gradient of PINNs.

To alleviate such problem, Wang et al. proposed the PirateNet structure [23]. With the residual adaptive operations, the model will be initialized as a single layer neural network. Thus, the initialization pathology induced by the increasing of the depth can be alleviated. Different from PirateNets, we aim to tackle the initialization pathologies by constructing a new neural network structure, which enables PINNs with strong nonlinear expressive ability even after the derivative operations.

## 3. Deeper-PINNs

The structure of Deeper-PINN is shown in Figure 1. The basic layer of Deeper-PINN is element-wise multiplied layer (EM-layer), which contains $k$ ($k$ > 1) sublayers. The inputs $\mathbf{x}$ are fed into $k$ fully connected layers with the same width. The outputs of the $k$ layers are then element-wise multiplied and get a non-linear high-dimensional output $\mathbf{H}^{(1)}$:

$$\mathbf{H}^{(1)} = \sigma\left(\mathbf{W}_1 \cdot \mathbf{x} + \mathbf{b}_1\right) \odot \sigma\left(\mathbf{W}_2 \cdot \mathbf{x} + \mathbf{b}_2\right) \odot \cdots \odot \sigma\left(\mathbf{W}_k \cdot \mathbf{x} + \mathbf{b}_k\right), \tag{8}$$

where $\sigma$ denotes activation function, $\mathbf{W}$ represents weights, $\mathbf{b}$ represents biases. $\mathbf{H}^{(1)}$ is then input into $N$ shortcut connection blocks. Each shortcut connection block contains 2 element-wise multiplication layers. With $\mathbf{H}^{(l)}$ as the input of $l$-th block ($1 \le l \le N$), the forward process of each shortcut connected element-wise multiplication block can be expressed as:

$$\mathbf{h}_1^{(l)} = \sigma\left(\mathbf{W}_{11}^{(l)} \cdot \mathbf{H}^{(l)} + \mathbf{b}_{11}^{(l)}\right) \odot \sigma\left(\mathbf{W}_{12}^{(l)} \cdot \mathbf{H}^{(l)} + \mathbf{b}_{12}^{(l)}\right) \odot \cdots \odot \sigma\left(\mathbf{W}_{1k}^{(l)} \cdot \mathbf{H}^{(l)} + \mathbf{b}_{1k}^{(l)}\right), \tag{9}$$

$$\mathbf{h}_2^{(l)} = \sigma\left(\mathbf{W}_{21}^{(l)} \cdot \mathbf{h}_{21}^{(l)} + \mathbf{b}_{21}^{(l)}\right) \odot \sigma\left(\mathbf{W}_{22}^{(l)} \cdot \mathbf{h}_{22}^{(l)} + \mathbf{b}_{22}^{(l)}\right) \odot \cdots \odot \sigma\left(\mathbf{W}_{2k}^{(l)} \cdot \mathbf{H}^{(l)} + \mathbf{b}_{2k}^{(l)}\right), \tag{10}$$

$$\mathbf{H}^{(l+1)} = \mathbf{h}_2^{(l)} + \mathbf{H}^{(l)}, \tag{11}$$

the final output $u_\theta$ is given by:

$$u_\theta = \mathbf{H}^{(N)} \cdot \mathbf{W}^{(N+1)} + \mathbf{b}^{(N+1)}. \tag{12}$$

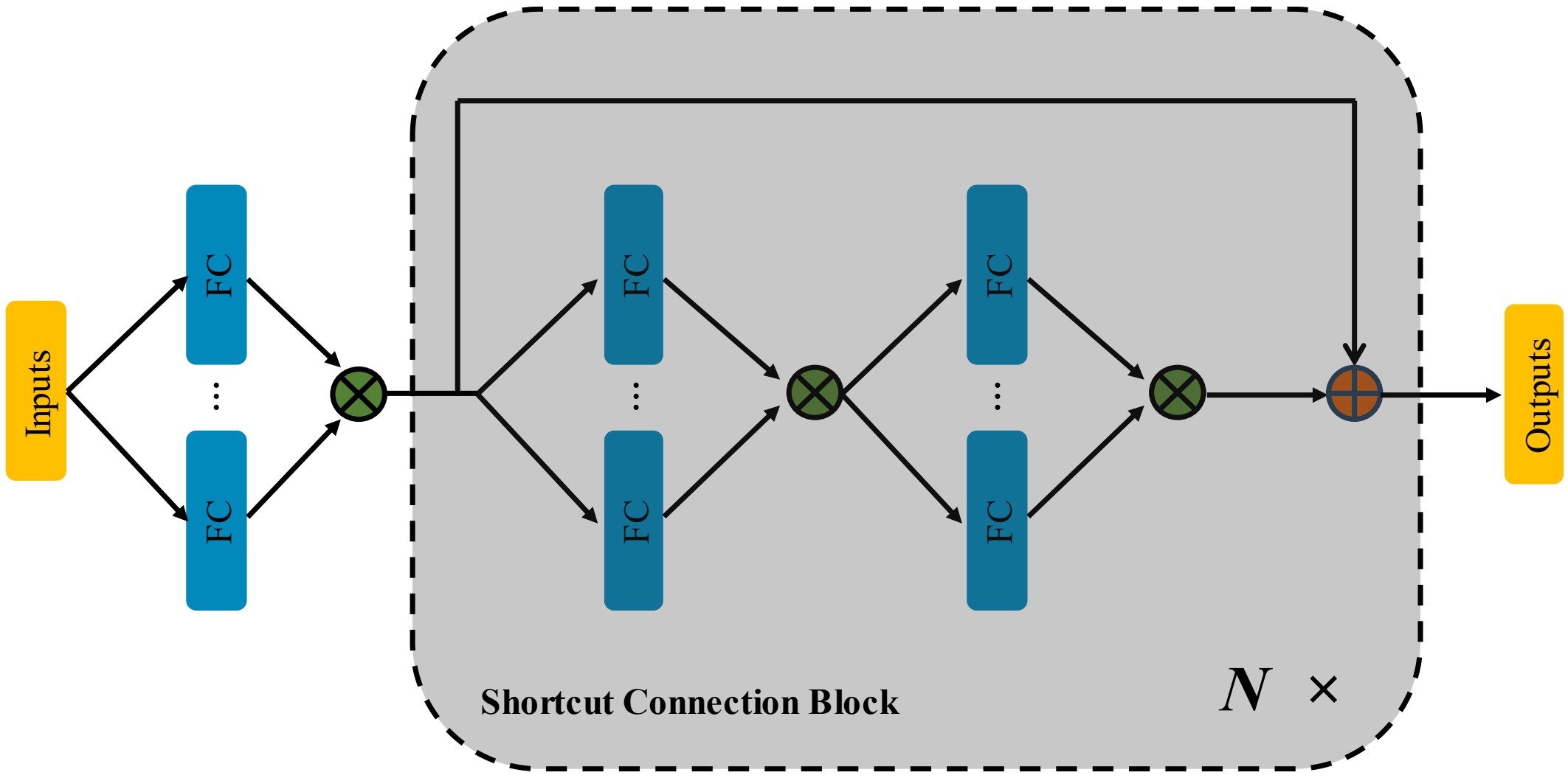

Figure 1: The structure of Deeper-PINNs.

Notice that the first layer of Deeper-PINN is responsible for transforming the dimension of the inputs. It can also be replaced by other structures such as a dense layer or random Fourier features.

In [23], Wang et al. verified the traditional MLP can suffer from the initialization pathology, which makes the neural network performs as a linear network. Under such circumstance, the model will be hard to approximate the derivatives of the solution. For our method, however, the element-wise multiplications can still project the features from different sublayers into a nonlinear space, which prevent the initialization pathology. The proof is shown below:

Consider the proposed structure without shortcut connections. With the input as $x$, the output of each hidden layer can be expressed as follows:

$$u_\theta^{(l)}(x) = g_1^l \odot g_2^l \odot \cdots \odot g_k^l,\ g_i^l = \sigma\left(\mathbf{W}_i^{(l)} u_\theta^{(l-1)}(x) + \mathbf{b}_i^{(l)}\right),\ l = 1, 2, \cdots L,\ i = 1, 2 \cdots k, \quad (13)$$

$L$ represents the number of hidden layers, $u_\theta^{(0)}(x) = x$. Assuming that all the layers of the neural network operate within a linear regime. Under such circumstance, $\sigma(x) \approx x$. When all biases are initialized as 0, the output of the final layer is:

$$u_\theta(x) \approx \prod_{i=1}^{L}\left(\left(\mathbf{W}_1^i \odot \mathbf{W}_2^i \odot \cdots \odot \mathbf{W}_k^i\right)^{L-i+1}\right) \cdot \mathbf{W}^{(L+1)} \cdot x^{k^L}. \quad (14)$$

The first-order derivate of $u_\theta(x)$ will be:

$$\frac{\partial u_\theta}{\partial x}(x) \approx k^L \cdot \prod_{i=1}^{L}\left(\left(\mathbf{W}_1^i \odot \mathbf{W}_2^i \odot \cdots \odot \mathbf{W}_k^i\right)^{L-i+1}\right) \cdot \mathbf{W}^{(L+1)} \cdot x^{k^L - 1}. \quad (15)$$

For MLP, as give in [23], the first-order derivate of $u_\theta(x)$ under such assumption is:

$$\frac{\partial u_\theta}{\partial x}(x) \approx \mathbf{W}^{(L+1)} \cdot \mathbf{W}^{(L)} \cdot \cdots \cdot \mathbf{W}^{(1)}. \quad (16)$$

Apparently, the MLP degrades as a deep linear network at initialization when calculating $\frac{\partial u_\theta}{\partial x}$. However, owing to the element-wise multiplication, our method still has nonlinear expressive ability.

To demonstrate the effectiveness of our method, we empirically validate our model on 1D advection equation which can be expressed as:

$$\frac{\partial u}{\partial t} + \beta \frac{\partial u}{\partial x} = 0,\ x \in [0,\ 2\pi],\ t \in [0,\ 1], \quad (17)$$

$$u(0,\ x) = \sin(x), \quad (18)$$

$$u(t,\ 0) = u(t,\ 2\pi), \quad (19)$$

where we set $\beta$ as 30. For Deeper-PINN, each layer contains 2 sublayers. The models are trained with Adam optimizer for 50,000 iterations with an initial learning rate of 0.001, and an exponential decay rate of 0.9 every 1000 iterations. All the outcomes are averaged from 5 independent trials.

Figure 2 shows the $L^2$ errors of ResNet [26] and the proposed method in different depth. It can be observed that when the depth of the neural network becomes larger, the performance of ResNet shows a degradation. However, benefiting from the element-wise multiplication operations, Deeper-PINN can obtain better results with the depth increasing.

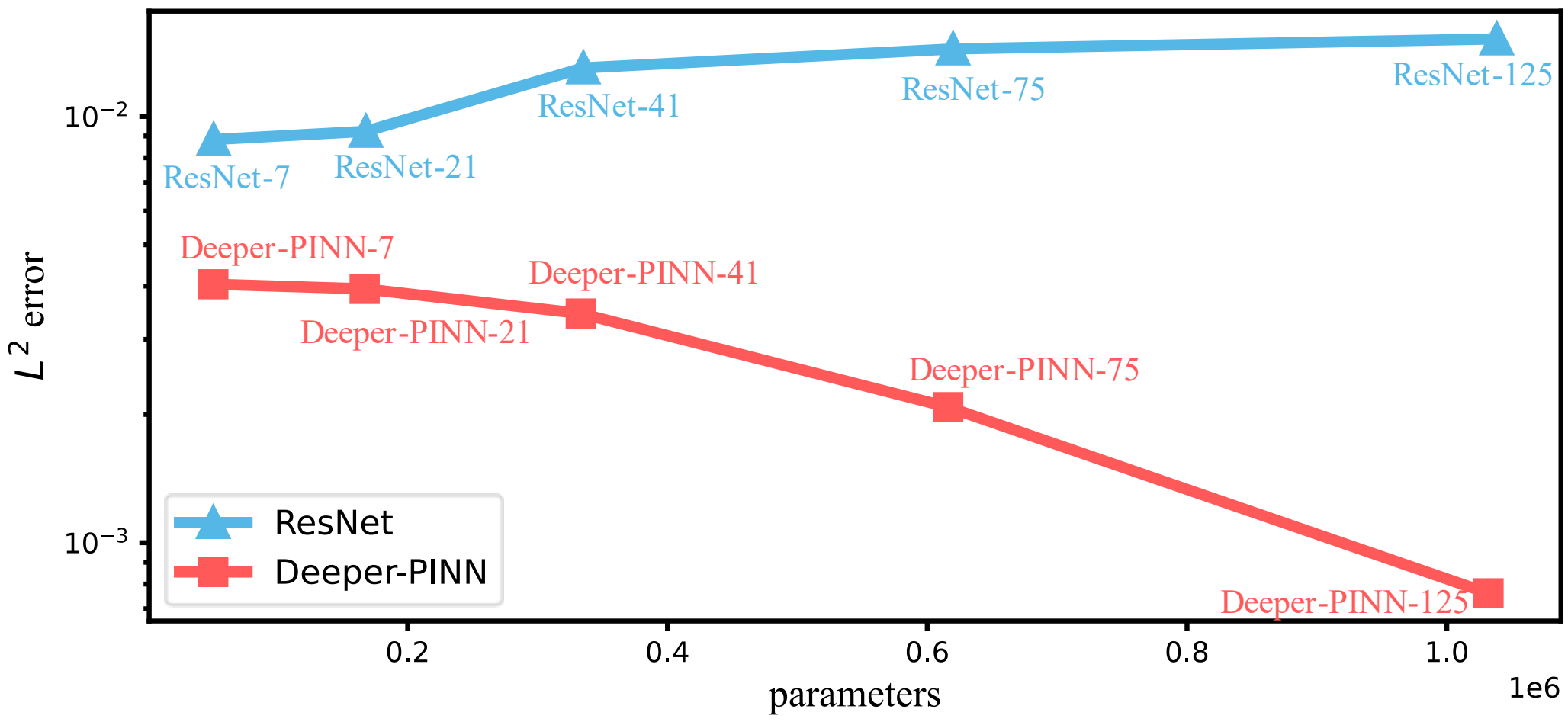

Figure 2: $L^2$ error of ResNet and Deeper-PINN in different depth.

# 4. Experiments

### Experimental setups

For all the cases, the weights are initialized with Xavier normal distribution [27], and tanh is used as the activation function. All the training points are resampled at every iteration to prevent the overfitting problem. Adam is utilised as the optimizer. The training is conducted on a single NVIDIA GeForce RTX 4090 GPU. All the results are averaged from 5 independent trials.

### Results

### Advection equation

We consider the 1D advection equation which is expressed as:

$$\frac{\partial u}{\partial t} + \beta \frac{\partial u}{\partial x} = 0, \quad (x,t) \in [0, 2\pi] \times [0,1], \tag{20}$$

$$u(0,\ x) = \sin(x), \tag{21}$$

$$u(t,\ 0) = u(t,\ 2\pi). \tag{22}$$

Previews studies have proofed that it is hard for PINNs to resolve advection equation when the transport velocity $\beta$ becomes larger. Here we set $\beta$ as 100, which has never been tried by PINNs.

Table 1: Advection equation: Hyper-parameter configuration.

| Parameter | Value |
|---|---|
| **PirateNet** | |
| Layer size | 100 |
| Fourier feature scale | 1.0 |
| **Deeper-PINN** | |
| Sublayer size | 100 |
| Number of sublayers per EM-layer | 2 |
| **Training Parameters** | |
| Initial learning rate | 0.001 |
| Exponential decay steps | 3,000 |
| Decay rate | 0.9 |
| Training steps | 120,000 |
| Collocation points | 5,000 |

To make it easier to resolve, the periodic boundary condition is directly imposed by a periodic layer [11]. EM-layer is adopted as the first hidden layer of Deeper-PINNs. More details of the hyper-parameter configuration can be found in Table 1.
The results and comparisons of PirateNets and Deeper-PINNs are shown in Figure 3. As can be seen in Figure 3 (a), the PirateNets fail to resolve this problem even with the layers increasing. For Deeper-PINNs, however, better results can be obtained with a deeper structure. Figure 3 (b) presents the training time of the 2 models. It can be observed that 45-layer Deeper-PINN takes about the same amount of training time with 58-layer PirateNet but has about 50% more parameters. This indicates that Deeper-PINNs have better computational efficiency.

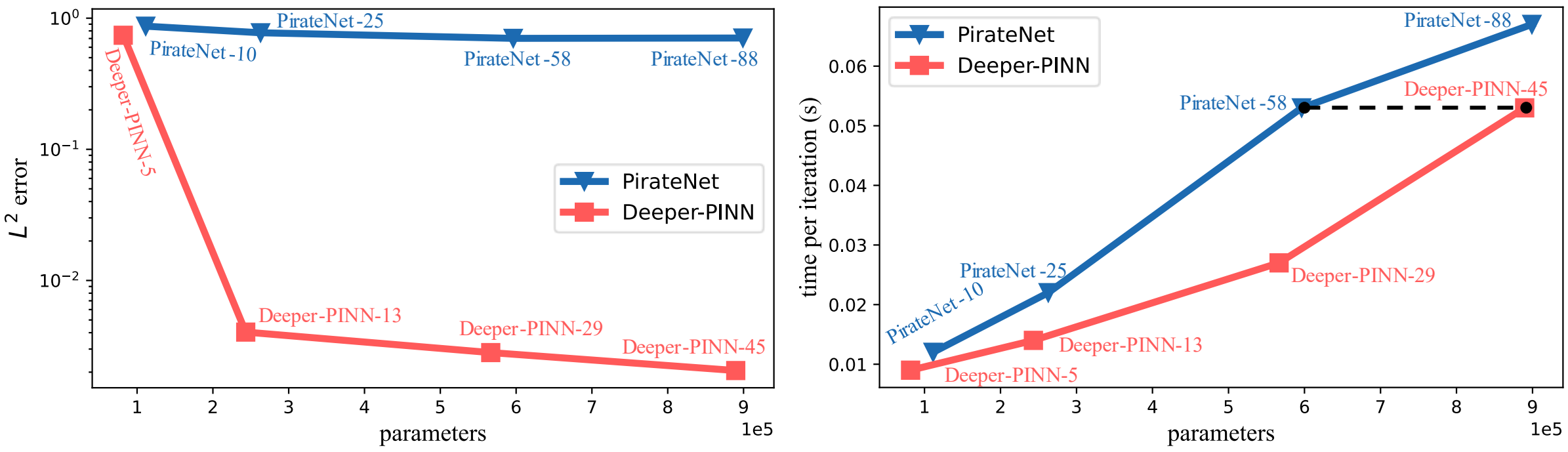

(a) (b)

Figure 3: Advection equation: Comparison between Deeper-PINNs and PirateNets. (a) $L^2$ error of PirateNet and Deeper-PINN in different depths. (b) Computing time of PirateNet and Deeper-PINN in different depths.

The loss and relative $L^2$ test error during the training of the 2 models are shown in Figure 4. The training process of Deeper-PINN-29 shows 2 distinct stages, which are fitting stage and diffusion stage [28]. For the fitting stage, the model tries to escape from the local minimum [29]. After that, the model moves into the diffusion stage, where most of the learning occurs. Correspondingly, at around 20000 iterations, the fitting stage ends and the $L^2$ test error decreases dramatically. However, for Deeper-PINN-5, PirateNet-10 and PirateNet-58, the training processes only show 1 stage, and the models fail to converge to correct solutions. This indicates that for Deeper-PINNs, a deeper structure enables the model to resolve complicated problems that a shallow structure cannot deal.

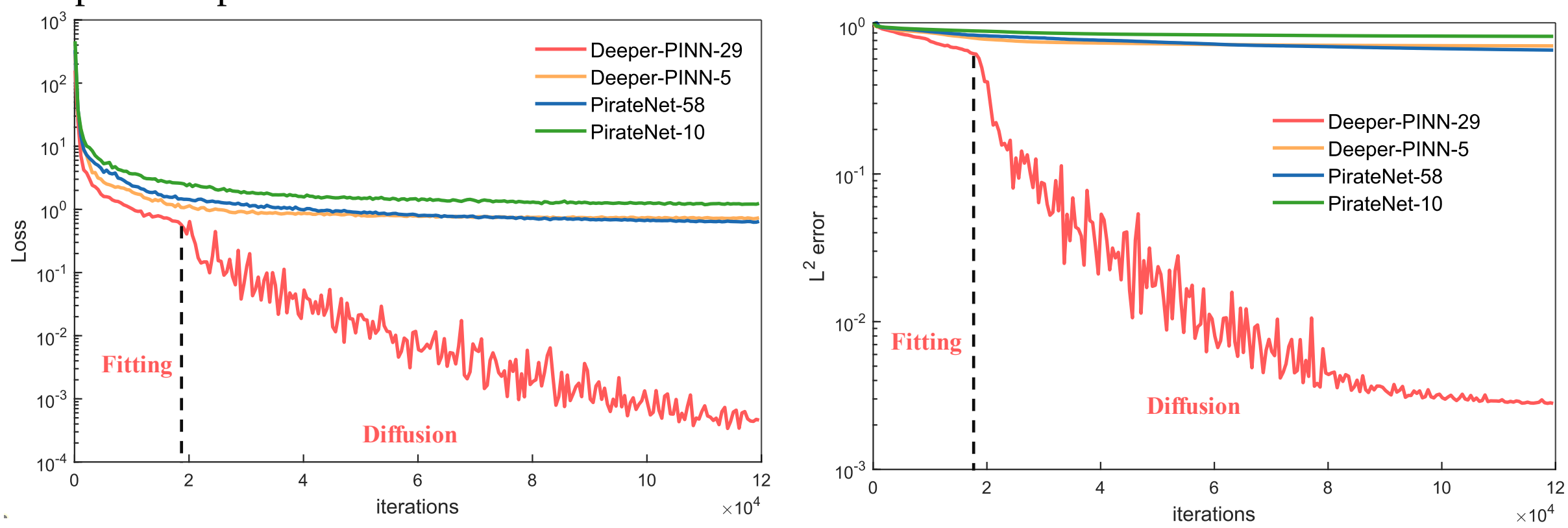

(a) (b)

Figure 4: Advection equation: (a) Loss curve during the training processes. (b) Relative $L^2$ test error during the training processes.

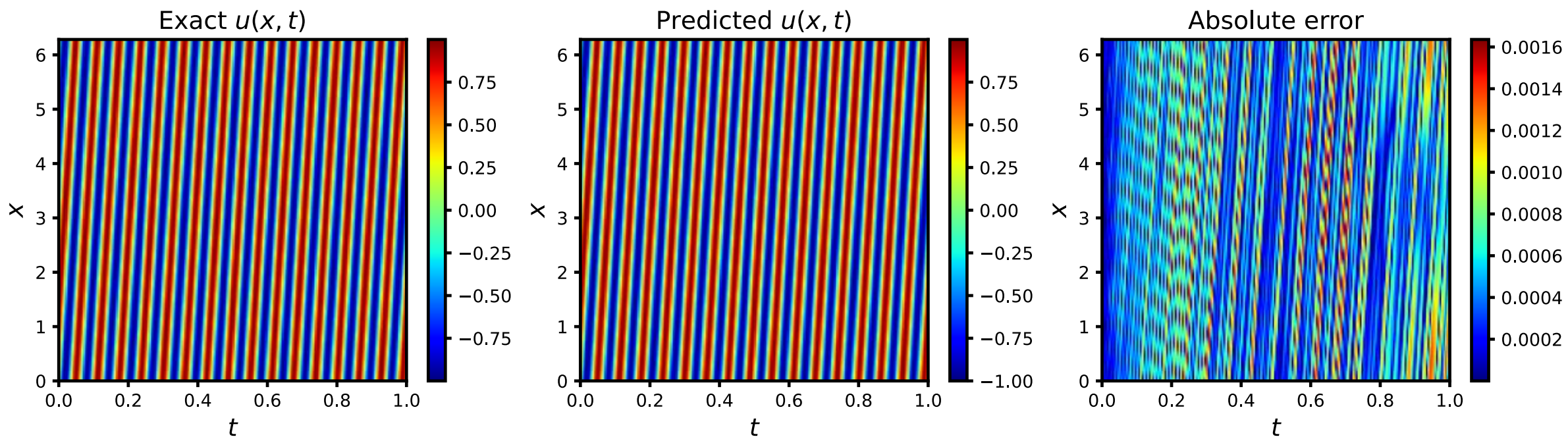

Figure 5. Advection equation: Comparison between prediction of Deeper-PINN and analytical solution. (a) $L^2$ error of PirateNet and Deeper-PINN in different depths. (b) Computing time of PirateNet and Deeper-PINN in different depths.

The best result of 45-layer Deeper-PINN is shown in Figure 5. With $L^2$ error of 8.76e-4, the prediction shows good alignment with the exact solution.

**Flow Mixing**

We consider flow mixing problem, which is expressed as below:

$$\frac{\partial u}{\partial t}+a\frac{\partial u}{\partial x}+b\frac{\partial u}{\partial y}=0,\ \ (x,y,t)\in[-4,4]\times[-4,4]\times[0,4] \tag{23}$$

$$a(x,y)=-\frac{v_t}{v_{t\max}}\frac{y}{r}, \tag{24}$$

$$b(x,y)=-\frac{v_t}{v_{t\max}}\frac{x}{r}, \tag{25}$$

$$v_t=\operatorname{sech}^2(r)\tanh(r), \tag{26}$$

$$r=\sqrt{x^2+y^2}, \tag{27}$$

The analytical solution is:

$$u(x,y,t)=-\tanh\left(\frac{y}{2}\cos(\omega t)-\frac{x}{2}\sin(\omega t)\right), \tag{28}$$

where $\omega=\frac{1}{r}\frac{v_t}{v_{t\max}}$. $v_{t\max}$ is set as 0.385.

Table 3: Flow mixing: Hyper-parameter configuration.

| Parameter | Value |
|---|---|
| **PirateNet** | |
| Layer size | 128 |
| Fourier feature scale | 2.0 |
| **Deeper-PINN** | |
| Sublayer size | 128 |
| Number of sublayers per EM-layer | 2 |
| Fourier feature scale | 2.0 |
| **Training Parameters** | |
| Initial learning rate | 0.001 |
| Training steps | 50,000 |
| Decay rate | 0.9 |
| Exponential decay steps | 2,000 |
| Collocation points | 20,000 |

The initial condition can be got from Equation.(28). Random Fourier features method is utilized as the first hidden layer to enhance the expressive ability of Deeper-PINN. The details of the hyper-parameter configuration can be found in Table 3.

The results are presented in Figure 8. Although PirateNet shows better performance at shallow structures, Deeper-PINN obtains smaller $L^2$ error when the neural networks become deeper. This indicates the superiority of Deeper-PINNs in deeper structures.

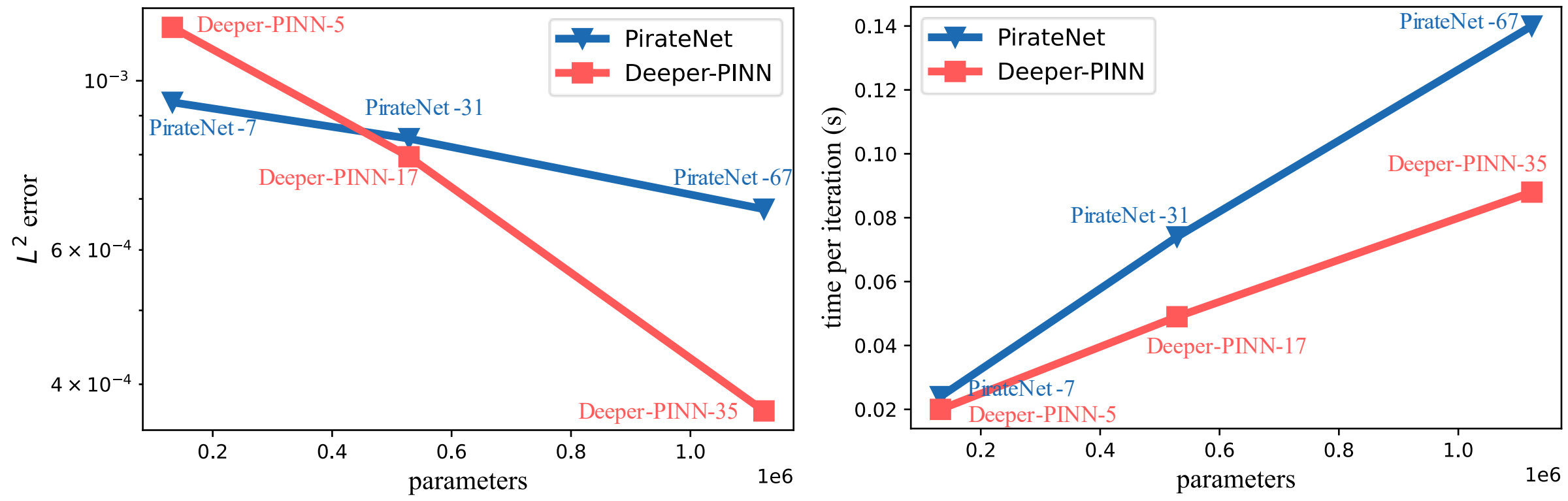

(a) (b)

Figure 8. Flow mixing: Comparison of the PirateNets and Deeper-PINNs. (a) $L^2$ error of PirateNet and Deeper-PINN in different depths. (b) Computing time of PirateNet and Deeper-PINN in different depths.

Figure 9 shows the predicted solution of PirateNet-35, with an $L^2$ error of 3.12e-4, the predicted solutions show good consistence with the exact solutions.

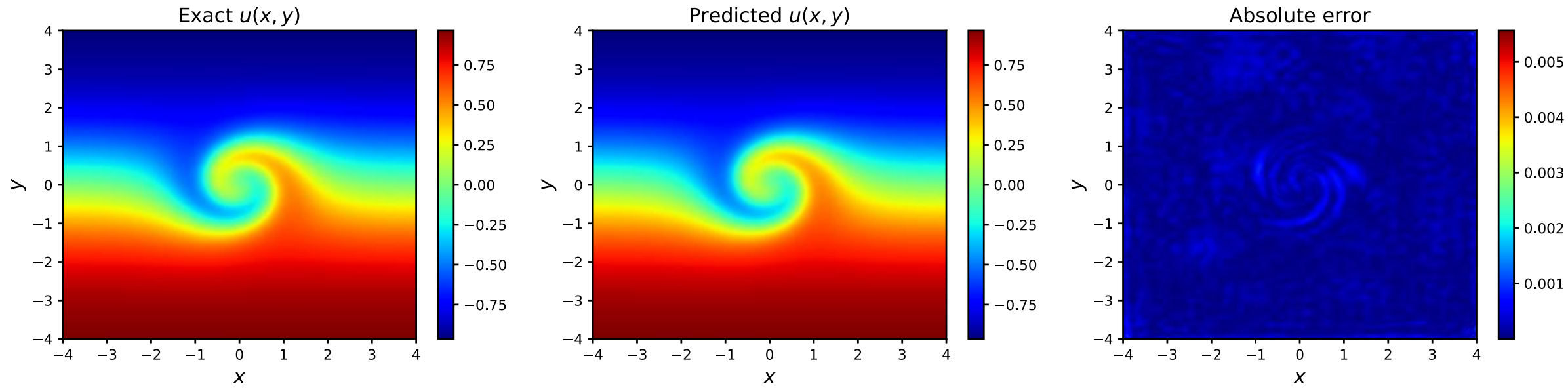

(a)

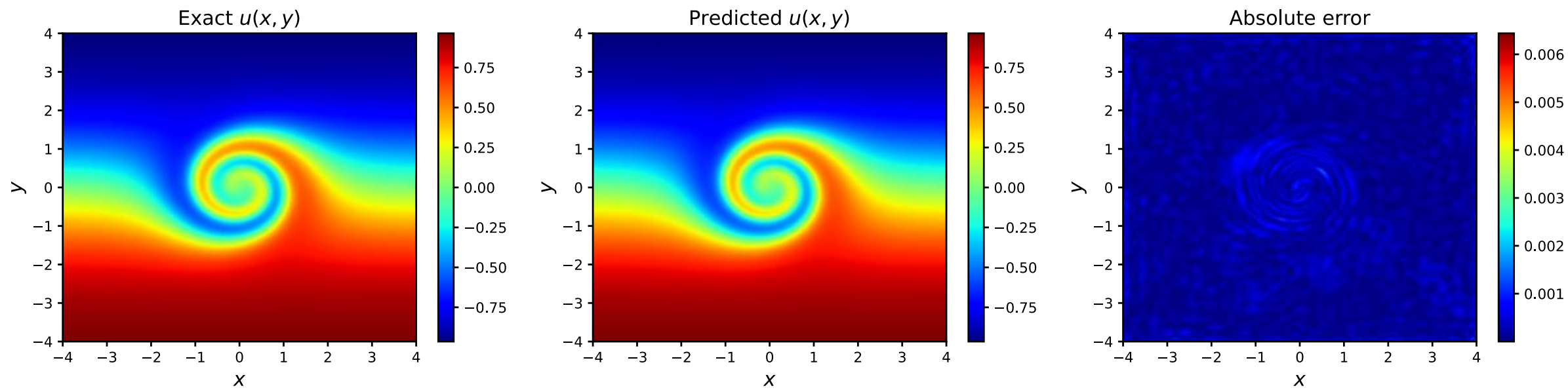

(b)

Figure 9. Flow mixing: Comparisons between prediction of Deeper-PINNs and analytical solutions (a) $t = 2$, (b) $t = 4$.

### 1D Wave propagation

We study the 1D wave propagation problem expressed as below:

$$\frac{\partial^2 u}{\partial^2 t} = c^2 \frac{\partial^2 u}{\partial x^2}, \quad (x,t) \in [0,1] \times [0,10], \tag{29}$$

$$u(0,t) = u(1,t) = 0,\ t \in [0,T], \tag{30}$$
$$u(x,0) = \sin(\pi x),\ x \in [0,1], \tag{31}$$
$$\frac{\partial u}{\partial t}(x,0) = 0, \tag{32}$$
where we set $c = 1$.The analytical solution for this problem is:
$$u(x,t) = \sin(\pi x)\cos(c\pi t). \tag{33}$$
Here the random Fourier features method is adopted as the first layer of Deeper-PINN. More details of the hyper-parameter configuration are listed in Table 2.

Table 2: 1D wave propagation: Hyper-parameter configuration.

| Parameter | Value |
|---|---|
| **PirateNet** | |
| Layer size | 100 |
| Fourier feature scale | 1.0 |
| **Deeper-PINN** | |
| Sublayer size | 100 |
| Number of sublayers per EM-layer | 3 |
| Fourier feature scale | 1.0 |
| **Training Parameters** | |
| Initial learning rate | 0.001 |
| Exponential decay steps | 500 |
| Decay rate | 0.9 |
| Training steps | 30,000 |
| Collocation points | 5,000 |

The comparisons of Deeper-PINNs and PirateNets are presented in Figure 6. Notice that for this problem, 2-layer Deeper-PINN achieves comparable $L^2$ error of 124-layer PirateNet with only about 5% parameters. The training time presented Figure 6 (b) indicates that Deeper-PINNs show better computational efficiency than PirateNets.

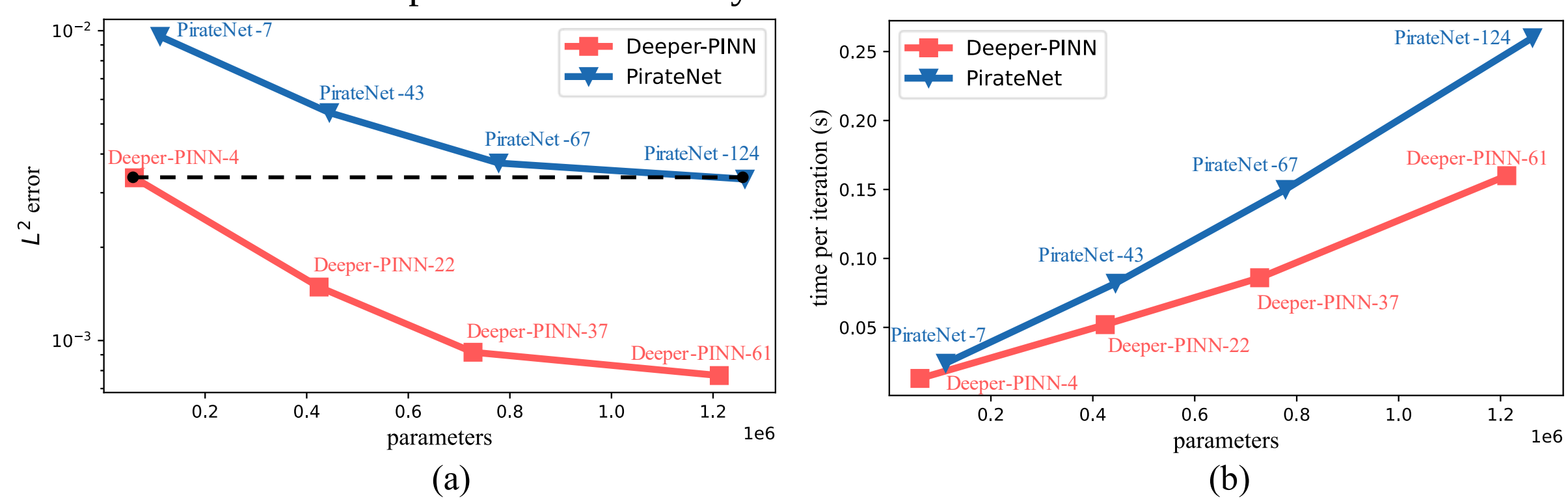

Figure 6. Wave propagation: Comparison between Deeper-PINNs and PirateNets. (a) $L^2$ error of PirateNet and Deeper-PINN in different depths. (b) Computing time of PirateNet and Deeper-PINN in different depths.

The best result of 64-layer Deeper-PINN shows an $L^2$ error of 4.54e-4. The corresponding prediction is shown in Figure 7.

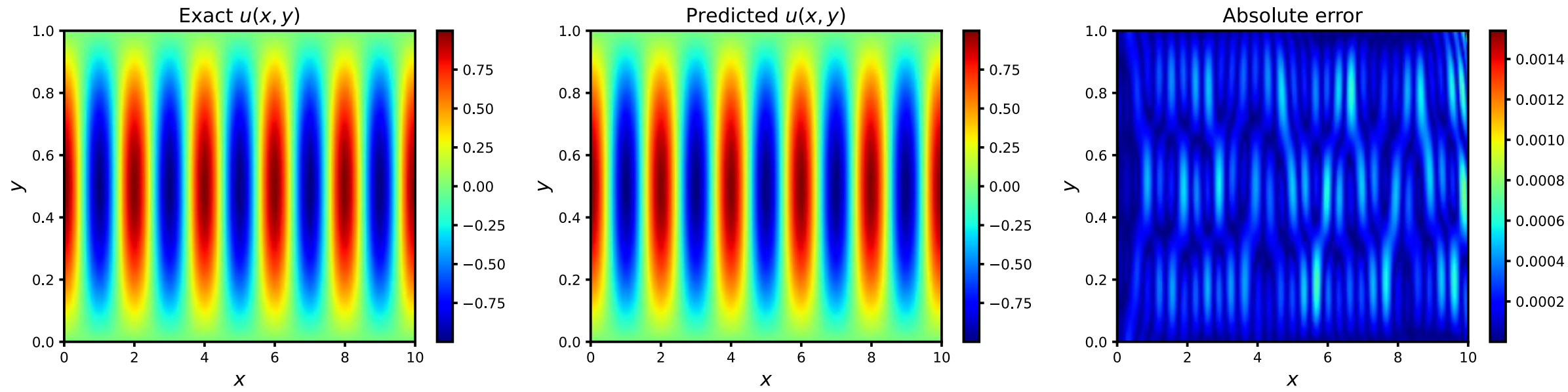

Figure 7. Wave propagation: Comparison between prediction of Deeper-PINN and exact solution.

**Korteweg–De Vries equation**

Korteweg–De Vries equation (KDV) expressed as below is considered.

$$u_t + \eta u u_x + \mu^2 u_{xxx} = 0,\ t \in (0,1),\ x \in (-1,1), \quad (34)$$

$$u(x,0) = \cos(\pi x), \quad (35)$$

$$u(t,-1) = u(t,1), \quad (36)$$

where we set $\eta = 1$ and $\mu = 0.022$.

The periodic boundary condition is directly imposed by a periodic layer. The random Fourier features method is adopted as the first hidden layer. More details of configuration can be found in Table 4.

Table 4: KDV: Hyper-parameter configuration.

| Parameter | Value |
|---|---|
| **PirateNet** | |
| Layer size | 128 |
| Fourier feature scale | 2.0 |
| **Deeper-PINN** | |
| Sublayer size | 128 |
| Number of sublayers per EM-layer | 2 |
| Fourier feature scale | 2.0 |
| **Training Parameters** | |
| Initial learning rate | 0.001 |
| Exponential decay steps | 2,000 |
| Decay rate | 0.9 |
| Training steps | 120,000 |
| Collocation points | 5,000 |

The results and comparison of Deeper-PINNs and PirateNets are presented in Figure 10. It can be observed that at shallow depth, these two models present similar accuracy. However, Deeper-PINNs show better performance when the neural networks become deeper. It's worht mentioning that with the depth increasing, the differences become more and more distinct as PirateNets appear to be saturated with the layers increasing. Also note that with about the same training time, the 35-layer Deeper-PINN gets about 50% smaller $L^2$ error than 43-layer PirateNet.

Figure 11 presents the best performance of 47-layer Deeper-PINNs, which shows good consistence with the exact solution. The $L^2$ error is 2.95e-4, which is better than the state-of-the-art of this case (4.27e-4) [23].

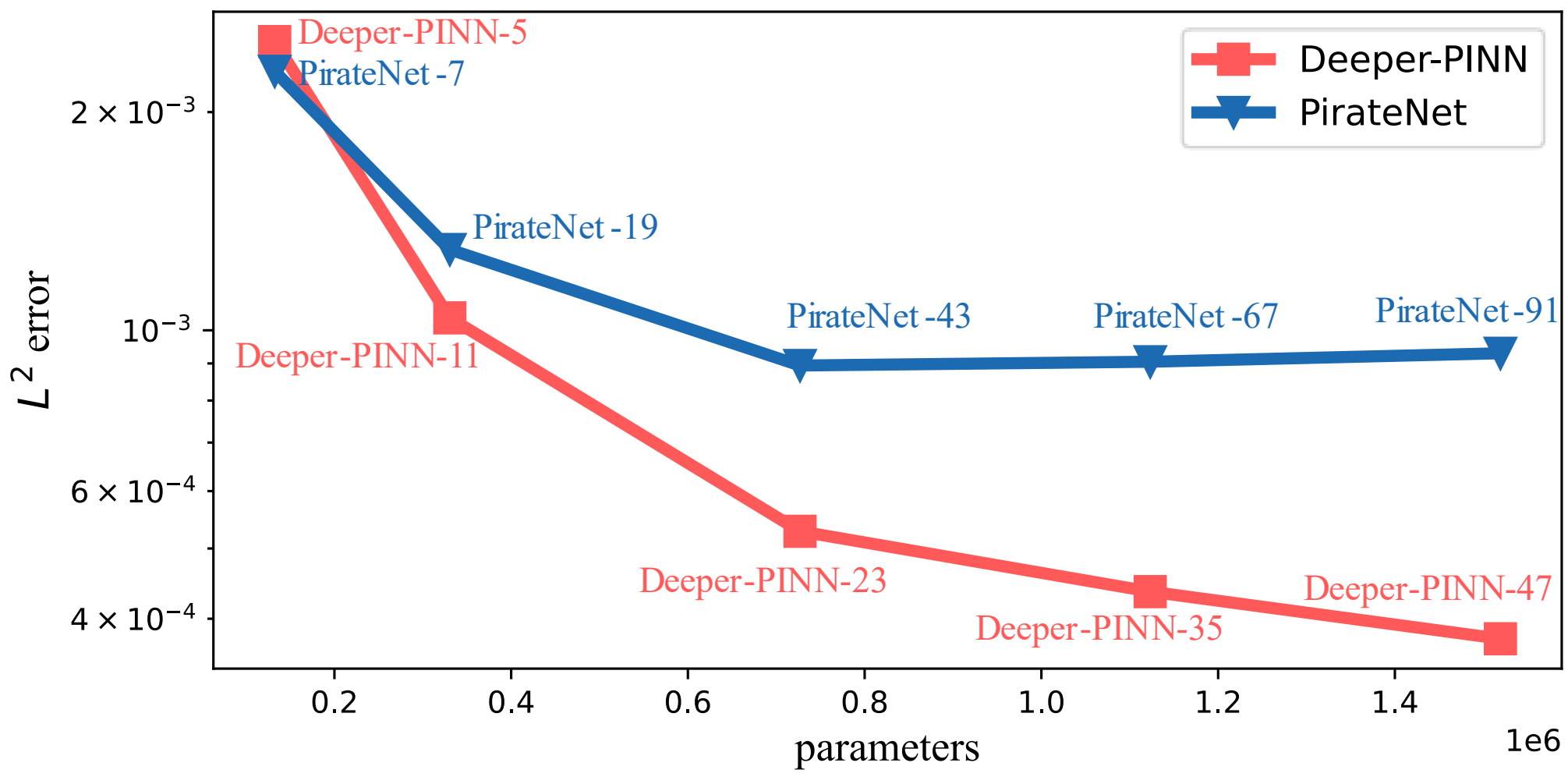

(a)

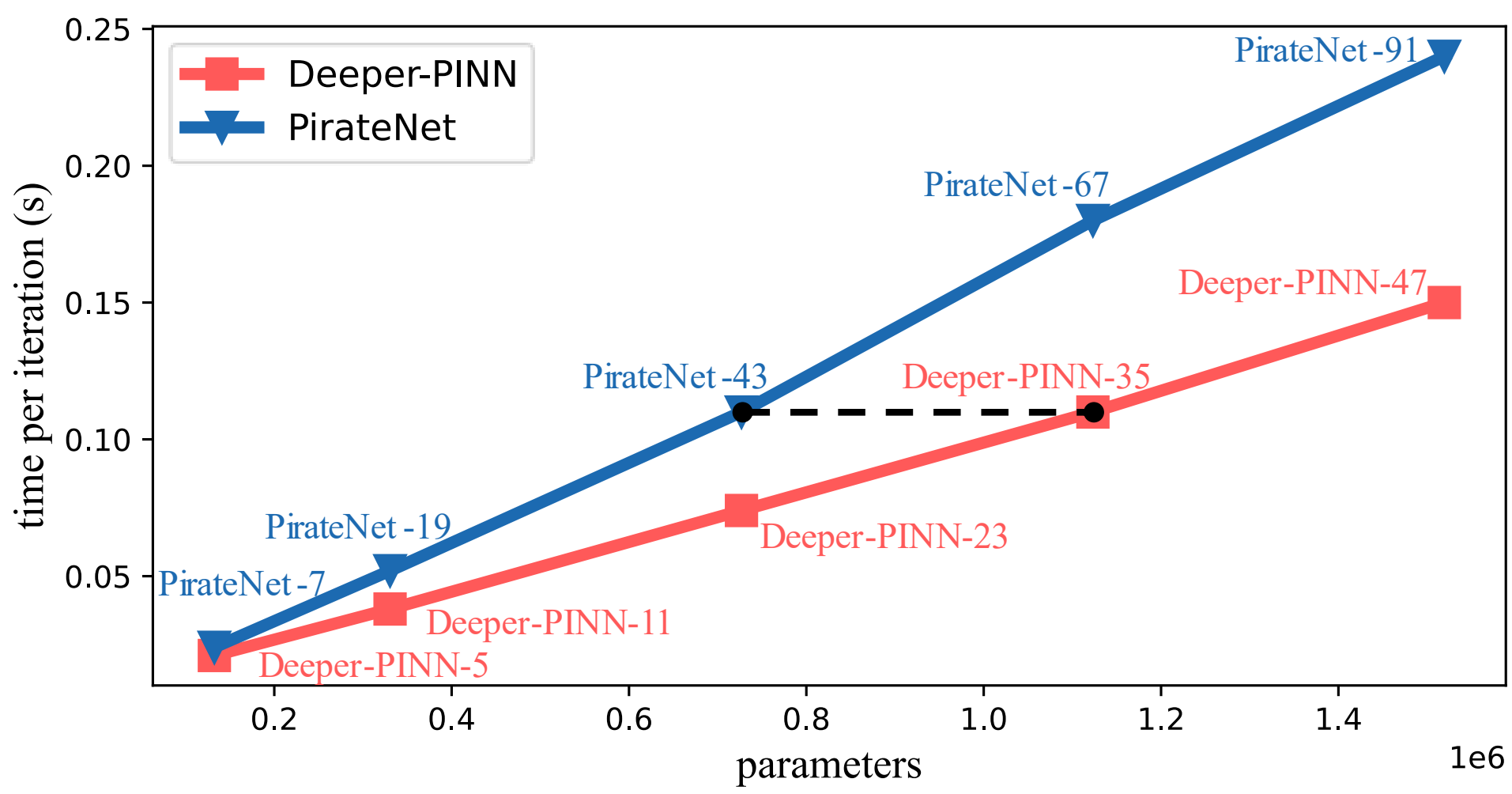

(b)

Figure 10. KDV: Comparison of the PirateNets and Deeper-PINNs. (a) $L^2$ error of PirateNet and Deeper-PINN in different depths. (b) Computing time of PirateNet and Deeper-PINN in different depths.

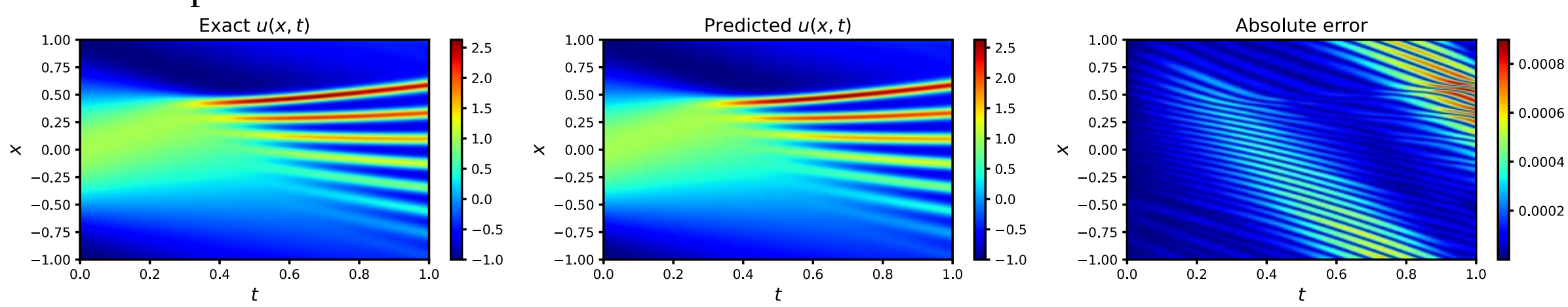

Figure 11. KDV: Comparison between prediction of Deeper-PINN and exact solution

# Conclusion

In this work, we proposed Deeper-PINNs, an element-wise multiplication based PINN structure. The element-wise multiplication operations could alleviate the initialization pathology of PINNs and allowed PINNs to go deeper. In our experiments, Deeper-PINNs showed better performance with the layers increasing. Moreover, compared with existing work, Deeper-PINNs showed better parameter efficiency and computing efficiency. In order to purely investigate the influence of the number of neural network layers on accuracy, we only used simple auxiliary method such as Random Fourier features and exactly imposed boundary conditions. We believe with more auxiliary methods such as loss balancing scheme, respecting causality and residual based attention etc., further improvements on accuracy of Deeper-PINNs can be gained.